\documentclass[letterpaper, 10 pt, conference]{ieeeconf}  %

\IEEEoverridecommandlockouts                              %

\usepackage{graphicx}
\usepackage{url}
\usepackage[caption=false]{subfig}
\usepackage[dvipsnames]{xcolor}
\usepackage{booktabs}
\usepackage{colortbl}
\usepackage{dashrule}
\usepackage{amsmath,amssymb}
\usepackage{arydshln}
\usepackage{dsfont}
\usepackage{mathtools}
\usepackage{microtype}
\usepackage[noadjust]{cite}

\newcommand\eg{\textit{e.g.}}
\newcommand\ie{\textit{i.e.}}

\DeclareMathOperator*{\argmin}{arg\,min}

\NewDocumentCommand\emojirobot{}{
    \includegraphics[trim=0cm 0.2cm 0cm 0cm,clip,height=0.65cm]{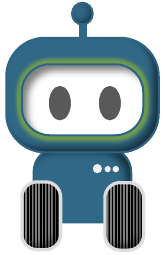}
}

\title{\LARGE \bf
Enhanced Model Robustness to Input Corruptions by \\ Per-corruption Adaptation of Normalization Statistics\emojirobot
}

\author{Elena Camuffo$^{1,3,*}$, Umberto Michieli$^{1}$, Simone Milani$^{3}$, Jijoong Moon$^{2}$, Mete Ozay$^{1}$
\thanks{*Research completed during internship at Samsung R\&D Institute UK.}%
\thanks{$^{1}$Samsung R\&D Institute UK (SRUK),
        Communications House, South St, Staines, Surrey, United Kingdom
        {\tt\small \{u.michieli,m.ozay\}@samsung.com}}%
\thanks{$^{2}$Samsung Research Korea,
        Seoul R\&D Campus, 56, Seongchon-gil, Seocho-gu, Seoul, Rep. of Korea
        {\tt\small jijoong.moon@samsung.com}}%
\thanks{$^{3}$University of Padova, via Gradenigo 6/b, 35129, Padova, Italy.
        {\tt\small \{elena.camuffo,simone.milani\}@dei.unipd.it}}%
}

\begin{document}

\definecolor{LightCyan}{rgb}{0.9,0.945,1}
\definecolor{LightYellow}{rgb}{0.99,0.955,0.84}

\definecolor{LightGrey}{rgb}{0.9,0.95,0.95}
\definecolor{LightRed}{rgb}{0.95,0.91,0.95}

\maketitle
\thispagestyle{empty}
\pagestyle{empty}

\begin{abstract}
Developing a reliable vision system is a fundamental challenge for robotic technologies (\eg, indoor service robots and outdoor autonomous robots) 
which can ensure reliable navigation even in challenging environments such as adverse weather conditions (\eg, fog, rain), poor lighting conditions (\eg, over/under exposure), or sensor degradation (\eg, blurring, noise), and can guarantee high performance in safety-critical functions.
Current solutions proposed to improve model robustness usually rely on generic data augmentation techniques or employ costly test-time adaptation methods. In addition, most approaches focus on addressing a single vision task (typically, image recognition) utilising synthetic data. 
In this paper, we introduce {Per-corruption Adaptation of Normalization statistics} (PAN) to enhance the model robustness of vision systems. Our approach entails three key components: (i) a corruption type identification module, (ii) dynamic adjustment of normalization layer statistics based on identified corruption type, and (iii) real-time update of these statistics according to input data. 
PAN can integrate seamlessly with any convolutional model for enhanced accuracy in several robot vision tasks. 
In our experiments, PAN obtains robust performance improvement on challenging real-world corrupted image datasets (\eg, OpenLoris, ExDark, ACDC), where most of the current solutions tend to fail. Moreover, PAN outperforms the baseline models by 20-30\% on synthetic benchmarks in object recognition tasks. 
\end{abstract}

\section{Introduction}
\label{sec:intro}
A reliable perception system is one of the key components of autonomous robotics, both for outdoor (\eg, autonomous driving systems) and indoor robotic systems (\eg, home service robots like smart vacuum cleaner robots). 

Advancements in deep learning technologies have led to the development of robust models for various robotic-related computer vision tasks, such as object recognition \cite{resnet,recht2019imagenet}, detection \cite{yolov8_ultralytics} and semantic segmentation \cite{Cordts2016Cityscapes,leak}. However, despite their high performance on standard benchmarks, these models often struggle with challenging environmental situations such as data corruptions \cite{hendrycks2019robustness}, adversarial attacks \cite{GoodfellowSS14}, and domain shifts \cite{HendrycksBMKWDD21}.
Factors like weather changes (\eg, \textit{snow}, \textit{frost}, \textit{fog}) or sensor degradation (\eg, \textit{shot noise}, \textit{defocus blur}) experienced by robotic systems can introduce natural alterations or data corruptions \cite{michieli2023online,Aakanksha023,barbato2024modular}. 
Moreover, the data-oriented nature of deep neural networks (DNNs) and their complex architectures make them vulnerable to even minor distribution shifts, resulting in significant performance degradation. To address these challenges, researchers have introduced datasets with synthetic corruptions \cite{hendrycks2019robustness,michaelis2020benchmarking}, and real-world data collections acquired in heterogeneous adverse conditions, both outdoors \cite{SDV19,SDV20,SDV21,Exdark} and indoors \cite{Bafghi_2023_CVPR,she2019openlorisobject}.
As robotic applications increasingly adopt deep learning models, equipping mobile autonomous robots %
with a robust vision system is of fundamental importance, to ensure reliable navigation in any environment and guarantee high-level performance even for safety-critical functions (\eg, autonomous driving, medical diagnostics, etc.).
A popular strategy to enhance model robustness is through data augmentation techniques during pre-training \cite{yucel2023hybridaugment,Chen_2021_ICCV,hendrycks2020augmix}, which aims to create a more generalizable model against corrupted images. 
Another approach is Test-Time Adaptation (TTA), which dynamically adjusts a pre-trained model's behaviour based on characteristics of  test data \cite{gong2022note,wang2021tent,niu2023towards}, enabling it to perform better in varying conditions at test time.

\begin{figure}[t]
    \centering
    \includegraphics[width=0.9\linewidth]{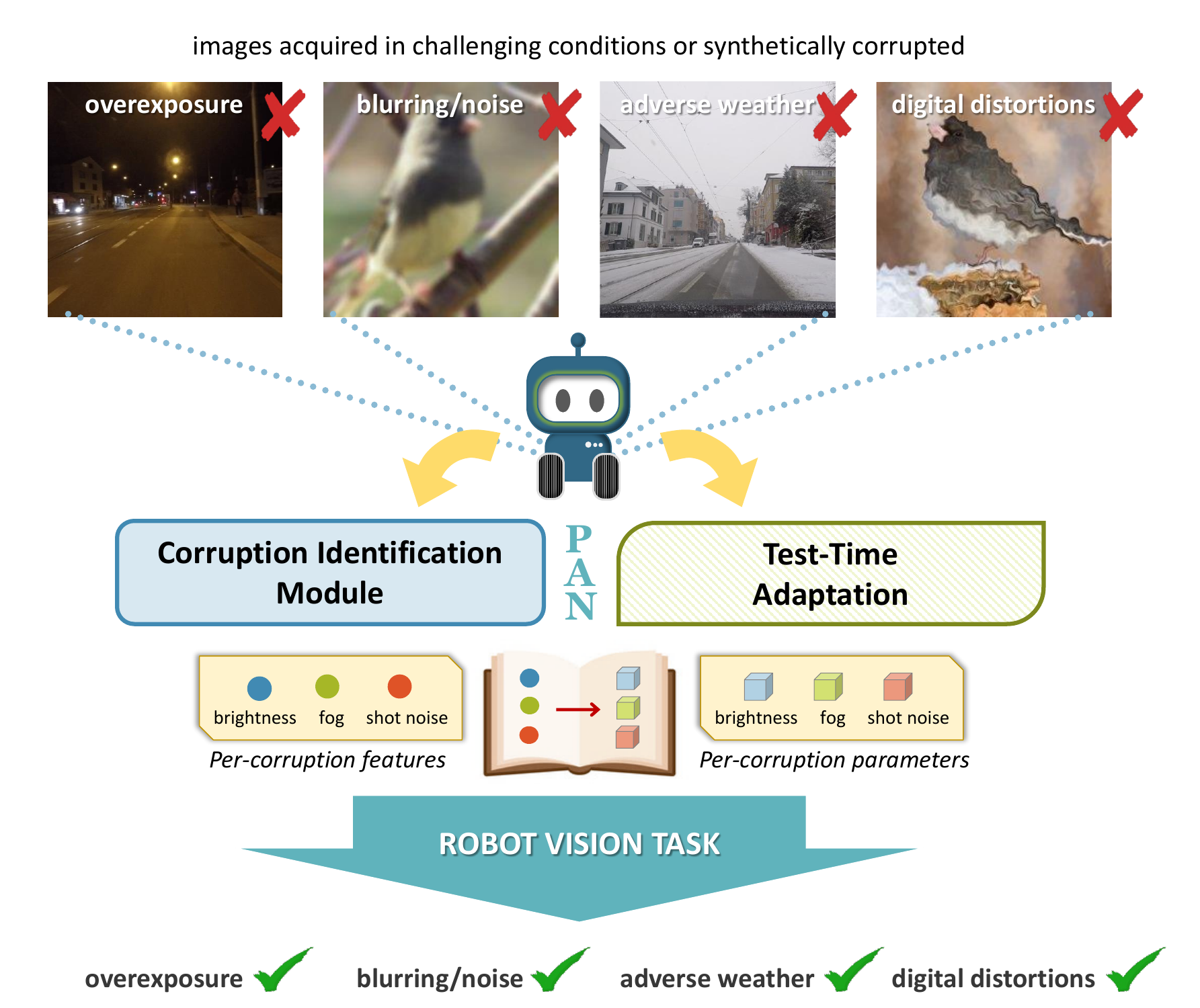} \vspace{-1em}
    \caption{Our approach enhances %
    robot vision systems via per-corruption adaptive normalization of neural network models. This is fundamental in challenging environmental situations with corrupted images. Our proposed PAN is built on (i) a corruption identification module (CIM) that extracts per-corruption features in order to recognize the input corruption, (ii) an inexpensive test-time adaptation step to adapt model parameters to the specific corruption type, (iii) a codebook to map features to candidate parameters.} 
    \label{fig:graph_abs}\vspace{-1em}
\end{figure}

In this paper, we propose PAN (\underline{P}er-corruption \underline{A}daptation of \underline{N}ormalization statistics), a novel strategy to improve model robustness in robotic applications (Fig.~\ref{fig:graph_abs}). PAN dynamically adapts normalization layers' parameters based on the type of corruption identified in an input image using a Corruption Identification Module (CIM). 
Our approach is simple yet effective, compatible with various convolutional architectures, and enhances accuracy on corrupted test data without burdensome training procedures. We demonstrate the effectiveness of PAN through extensive evaluations, achieving up to 30\% relative accuracy gain compared to the state of the art. Our method is computationally and memory efficient, hence it is suitable for on-device robotic applications. 

\section{Related Work}
\label{sec:rel}
Common image corruptions have various causes and occur frequently in real-world situations. These issues can be problematic for a wide variety of robot-related tasks, including localization \cite{rada}, navigation \cite{HAIDER20229060} and vision-related tasks \cite{ida,michieli2023online} (also dealt with in this work).
Considering robot vision, mainstream methods for tackling this problem can be broadly distinguished into two main categories: data augmentation and test-time adaptation methods.

\textbf{Data Augmentation}
methods improve model performance during pre-training via data augmentation: they aim to develop a general robust model against corrupted images \cite{Hendrycks2021PixMixDP,hendrycks2020augmix,Chen_2021_ICCV,yucel2023hybridaugment}. Some methods improve the robustness of models by automatically searching for improved data augmentation policies among common methods \cite{autoaugment}, or applying random noise or patches to train images \cite{Rusak2020ASW,lopes2019improving}. Other approaches transform each image in a dataset by mixing it with a collection of images \cite{hendrycks2020augmix,wang2021augmax} or automatically generating patterns \cite{Hendrycks2021PixMixDP}, to improve model generalization by out-of-distribution examples and prevent overfitting on the training distribution. This technique is effective also on robot vision tasks other than image recognition \cite{ida}. 
Recent works propose mixed augmentation strategies in the frequency domain \cite{Camuffo2023d}, as common corruptions mostly affect frequency components: APR \cite{Chen_2021_ICCV} re-combines the phase spectrum of one image and the amplitude spectrum of another image, HA \cite{yucel2023hybridaugment} includes hierarchical augmentations at variable frequency spectra.

\textbf{Test-Time Adaptation (TTA)} methods focus on resolving data distribution shift at test-time, using data from target domains \cite{gong2022note,wang2021tent,lame}. These methods have been widely employed in robotic-related vision problems such as image registration \cite{zhu2021testtime}, depth estimation \cite{tta_s2r} and point cloud upsampling via meta-learning \cite{hatem2023testtime}.
There are different types of TTA methods. Some assume that target domain data can be observed simultaneously during adaptation, \eg, adapting a source model on the target domain data using self-supervised loss, and employing the features obtained from the intermediate layers of the adapted model to refine the pseudo labels for the entire target dataset \cite{Lee2013PseudoLabelT}.
Other TTA methods assume that target data is received by the system in mini-batches \cite{zhang2022memo,lame} and 
updates statistics at each iteration. For instance, AugBN \cite{Khurana2021SITASI} estimates normalization statistics of the unseen test distribution from the given test images in a mini-batch, using only one forward pass.
TTA can also be applied to streams of data (sampled from a new data distribution, distinct from the source data distribution) instead of a fixed test set, in an online manner \cite{wang2021tent}. However, samples obtained at test time may come from a variety of different distributions, leading to new challenges, such as error accumulation and catastrophic forgetting  \cite{Camuffo_2023_CVPR}. To address this issue, a few methods \cite{wang2022continual,niu2022efficient} investigate the continual test time adaptation problem that adapts the pre-trained source model to the continually changing test data. SAR \cite{niu2023towards} employs an optimization scheme, which removes samples with large gradients and encourages model weights to lie in a flat minimum. NOTE \cite{gong2022note} performs a selective mixing strategy that only calibrates the batch normalization layers statistics for detected out-of-distribution samples. ONDA \cite{onda} estimates BN statistics via running average on test batches.

\begin{figure}[t]
    \centering
    \includegraphics[trim=2cm 0cm 8.4cm 0cm,clip,height=4.2cm]{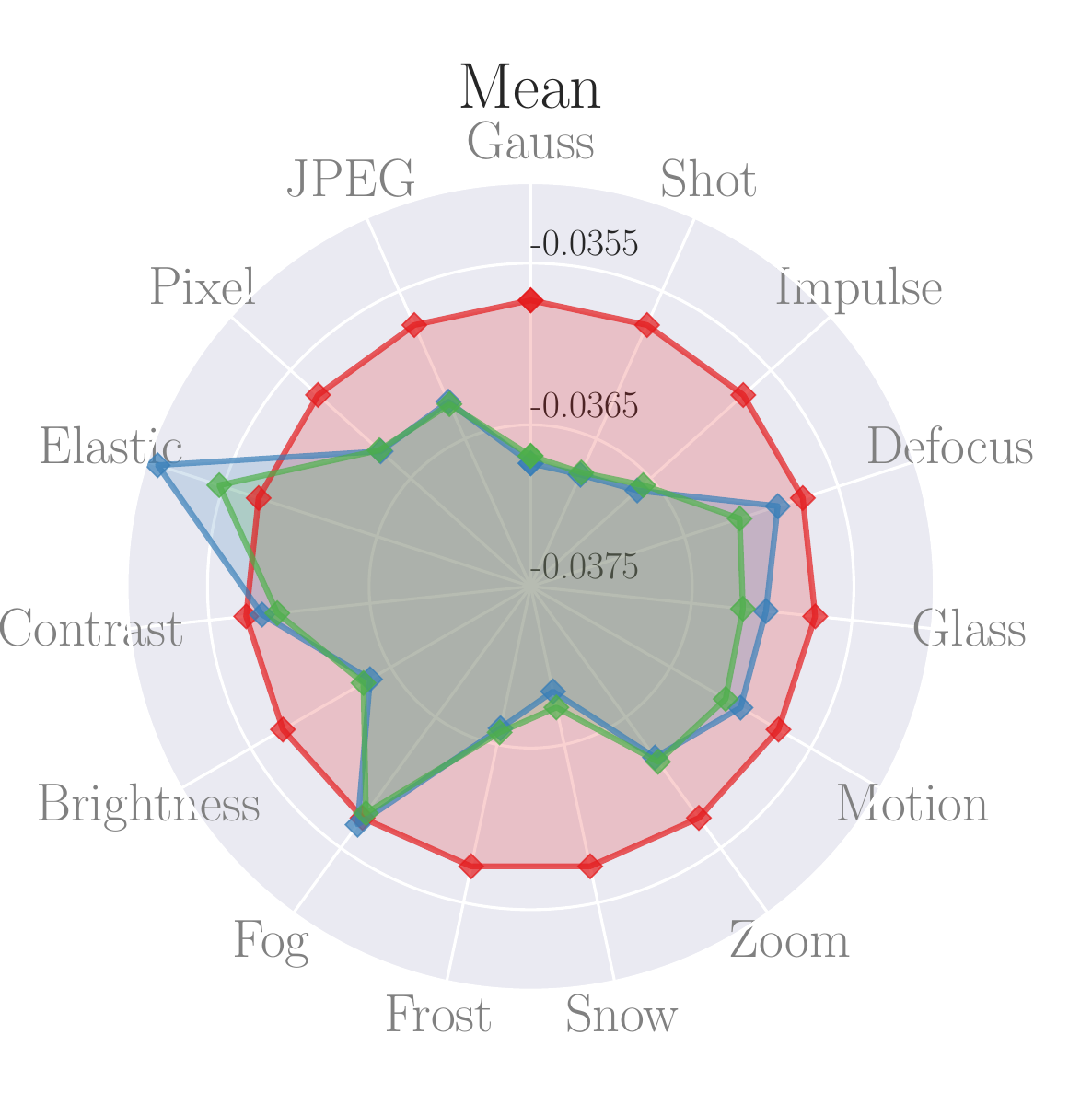}
    \hfill
    \includegraphics[trim=5.5cm 0cm 5.6cm 0cm,clip,height=4.2cm]{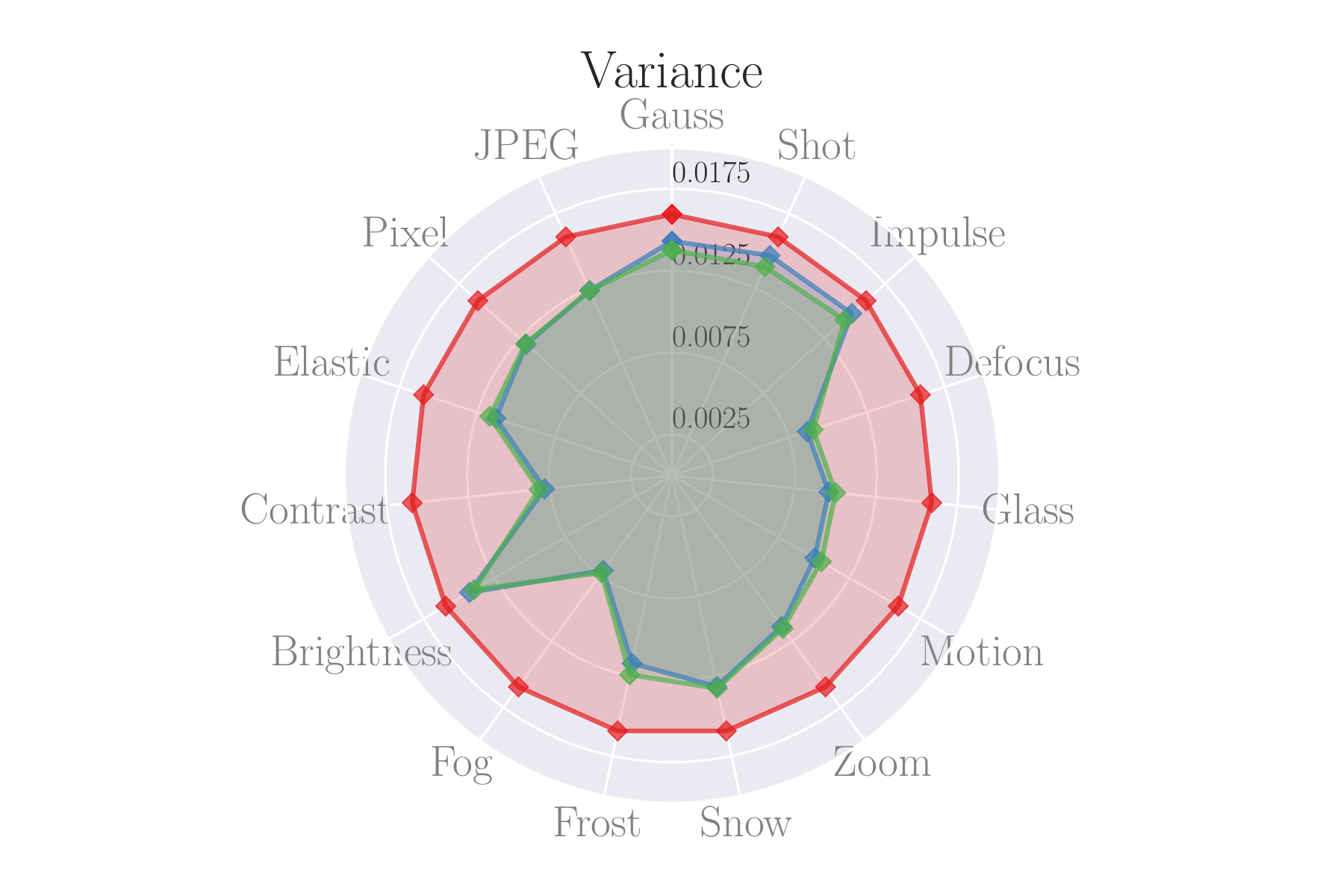} 
    \vskip -4.5ex
    \includegraphics[trim=22cm 12cm 0.6cm 3cm,clip,width=1.7cm]{figs/spidermean.pdf} \vskip -2ex
    \caption{Statistics estimated at normalization layers vary depending on the image corruption type, {averaged on all layers (ResNet18 on ImageNet-C)}. Unlike classical data augmentation approaches where a single set of normalization statistics is estimated for all corruption types on a source domain (\textcolor{Red}{\textbf{red}}), our method estimates normalization statistics for each corruption (\textcolor{NavyBlue}{\textbf{blue}}), which are very close to the reference ones, estimated assuming that the true corruption type of the data is known (\textcolor{ForestGreen}{\textbf{green}}).}
    \label{fig:bn_layers}
\end{figure}

\begin{figure}
    \centering
    {\scriptsize {\hspace{0.2cm} Mean \hspace{3.3cm} Variance}\\
    \includegraphics[trim=0cm 0cm 0cm 1.4cm,clip,width=1.02\linewidth]
    {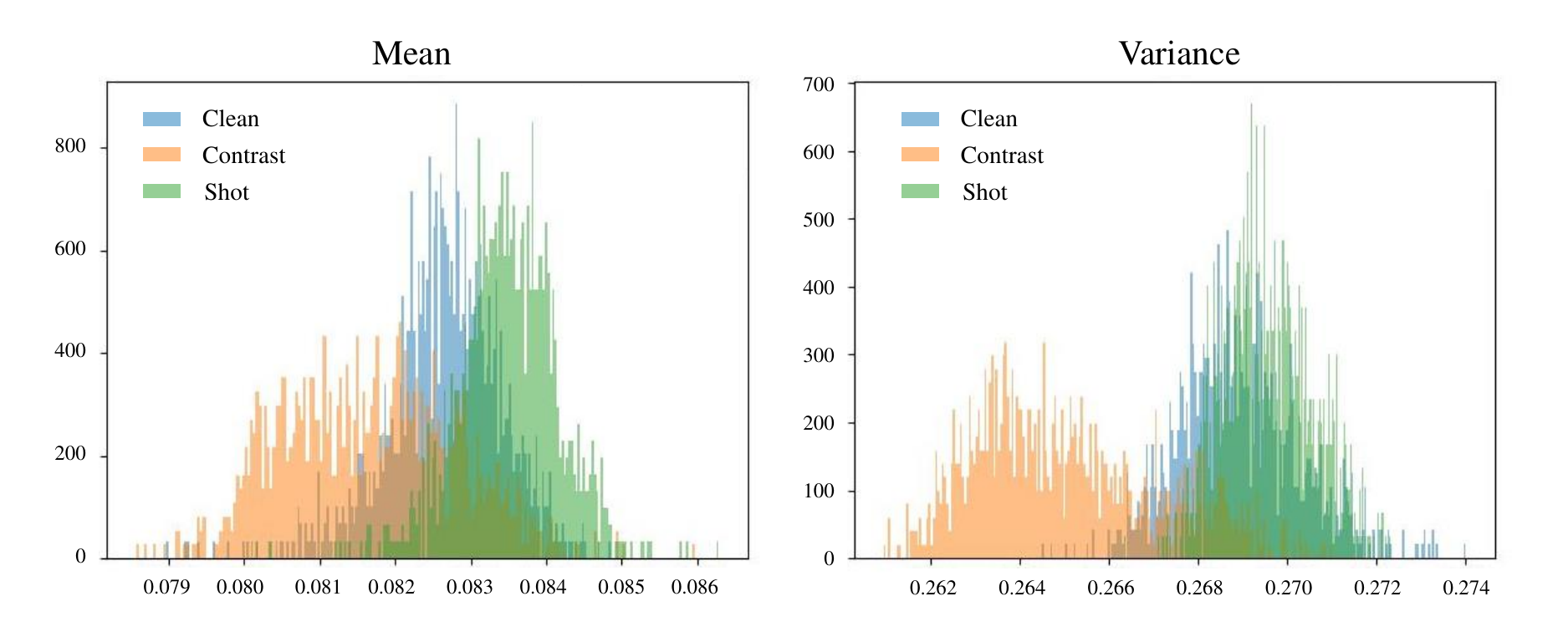}} \vskip -2ex
    \caption{Mean and variance distributions of the output of the first BN layer when encountering \textit{clean} data, \textit{contrast} corrupted data and \textit{shot} noise corrupted data (ResNet18 on ImageNet-C). 
    }
    \label{fig:hist} \vspace{-1em}
\end{figure}

\begin{figure*}[t]
    \centering
     \includegraphics[width=0.78\linewidth]{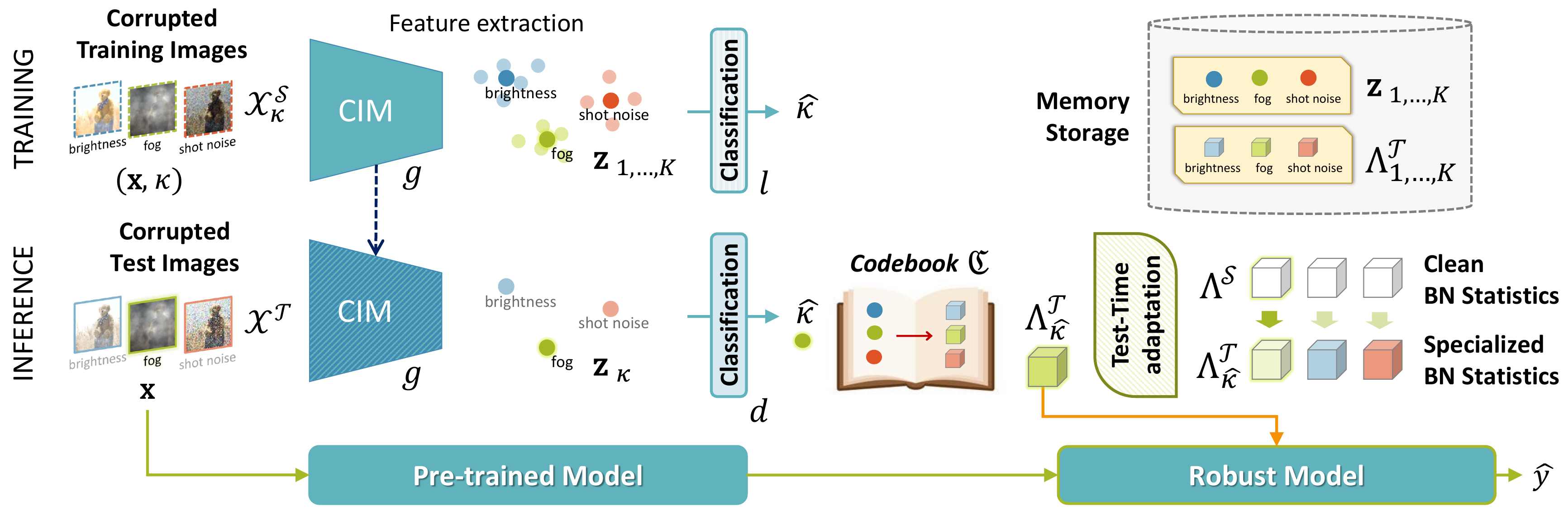}\vspace{-1em}
    \caption{The corruption identification module (Sec.~\ref{sec:corrid}) is trained on corrupted training images and a set of corruption-related prototypical features $\bar{\mathbf{z}_{1,\dots,K}}$ is built, by averaging features $\mathbf{z}$ relative to each corruption. Then, at \textit{inference time}, the CIM is frozen and a \textit{Codebook} $\mathfrak{C}$ (Sec.~\ref{sec:codebook}) maps the corruption identified by the CIM to the respective corruption-specific BN parameters. Such parameters are initialized with the ones of the pre-trained downstream task model $F(\cdot)$ on clean source images $\mathcal{X}^{\mathcal{S}}$ and adapted to test images via TTA, separately for each identified corruption $\hat{\kappa}$
    (Sec.~\ref{sec:specialization}), obtaining a corruption-specific set $\Lambda_{\hat{\kappa}}^{\mathcal{T}}$. Finally, $\Lambda_{\hat{\kappa}}^{\mathcal{T}}$ is plugged into $F(\cdot)$ achieving enhanced robustness on downstream tasks, specifically on the identified corruption. {The systems stores $\mathbf{z}_{1,\dots,K}$ and $\Lambda^{\mathcal{T}}_{1,\dots,K}$ to use and update them while doing inference.}} \vspace{-1em}
    \label{fig:pipeline}
\end{figure*}

\section{Methodology: Per-corruption Adaptive Normalization (PAN)}
\label{sec:method}
Our approach builds upon the observation that statistics of BN layers in any convolutional architecture significantly differ for images corrupted according to different corruption types (Fig.~\ref{fig:bn_layers}), but are similar for images with the same corruption type (Fig.~\ref{fig:hist}).
Some previous work \cite{Nado2020EvaluatingPB,Schneider2020ImprovingRA,gong2022note} explored adaptation of statistics of normalization layers for TTA, keeping a single set of normalization parameters for all corruptions, to build generic normalization layers to accommodate any input corruption. 
Instead, we build multiple sets of normalization statistics estimated for each corruption type. PAN is composed of three parts:
\begin{enumerate}
\item A corruption type identification module (Sec.~\ref{sec:corrid}). 
\item  A per-corruption adaptation method for adapting statistics of BN layers to various corruption types at inference time (Sec.~\ref{sec:specialization}). 
\item A codebook to map the identified corruption type to the respective set of BN statistics (Sec.~\ref{sec:codebook}).
\end{enumerate}

\subsection{Problem Setup: Improving Model Robustness} 

\textbf{Image corruption:}
Let $F( \mathbf{x}, y;\mathcal{W})$ be a DNN model mounted on a robot for visual scene understanding. The aim of $F(\cdot)$ is to approximate ground truth labels $y \in \mathcal{Y}$ of input images ${\mathbf{x} \in \mathcal{X} \subset \mathbb{R}^{w \times h \times 3}}$ optimizing its set of learnable parameters $\mathcal{W}$ (\eg, weights and biases of the network architecture of the model). Among these parameters, we denote the set of parameters of its BN layers by $\Lambda \subset \mathcal{W}$. 
Samples of a source (\textit{clean}) dataset ${\mathcal{D}^{\mathcal{S}} = \{\mathcal{X}^{\mathcal{S}}, \mathcal{Y}^{\mathcal{S}}\}}$ are drawn from a probability distribution $P^\mathcal{S}(\mathbf{x})$ on a \textit{source} domain $\mathcal{S}$.
Then, we consider a target (\textit{corrupted}) dataset ${\mathcal{D}^{\mathcal{T}} = \{\mathcal{X}^{\mathcal{T}}, \mathcal{Y}^{\mathcal{T}}\}}$ of distorted images sampled from a \textit{target} domain $\mathcal{T}$. We make a distinction between real (\textit{endogenous}) and synthetic (\textit{exogenous}) distortions as follows:

\textbf{Endogenous distortions} are natural corruptions that imply a shift in image statistics due to either inherent noise of camera sensors, deformations of objects observed in the images, or divergence of patterns of the objects.
This is the most general case, where \textit{target} test data $\mathcal{X}^{\mathcal{T}}$ cannot be parametrized by any operator. We denote a corrupted set of images presenting the same type of corruption, \eg, dark images, as {{$\mathcal{X}_{u}^{\mathcal{T}} \sim P_u^\mathcal{T}(\mathbf{x})$}}, where $u$ denotes the corruption type.
The distribution of the images in the corrupted set is different from the source $P^\mathcal{S}( \mathbf{x})$.

\textbf{Exogenous distortions} {are synthetic approximations of real corruptions provided by a function of clean images.
They are obtained assuming that there exists an operator $C_{k,s}$ which corrupts a given set of clean images $\mathcal{X}^{\mathcal{S}}$ by ${C_{k,s} (\mathcal{X}^{\mathcal{S}}) \eqcolon \mathcal{X}_{k,s}^{\mathcal{S}}}$ where $k$
denotes the synthetic corruption type and $s$ denotes its severity level.
Images of each corrupted set $\mathcal{X}_{k,s}^{\mathcal{S}}$ are sampled from 
${P_{k,s}^\mathcal{S}( \mathbf{x}) = \psi_{k,s}(P^\mathcal{S}( \mathbf{x})})$ as the operator $C_{k,s}$ transforms the distribution by a non-linear transformation $\psi_{k,s}(\cdot)$ according to the corresponding corruption type $k$ and severity $s$. 
Synthetic corruptions attempt to approximate real corruptions, \ie, $\psi_{k,s}(P^\mathcal{S}( \mathbf{x})) \approx P_u^\mathcal{T}(\mathbf{x})$ for some $u$.}

\subsection{Corruption Identification Module (CIM)}
\label{sec:corrid}
Our CIM is designed using a convolutional encoder followed by a linear classifier, {taken} from \cite{millerclass}. 

\textbf{Architecture of the CIM.} Extraction of corruption-specific features is accomplished through a DNN model $r=l \circ g$ \cite{millerclass}, composed of a convolutional encoder ${g}(\cdot)$ that projects an input image $\mathbf{x}$ to a feature vector by $\mathbf{z} = g(\mathbf{x})$, and a linear layer $l(\mathbf{z})$ that outputs corruption identification probabilities. 

\textbf{Training:}
{The CIM performs a corruption classification task to recognize the corruption type of input images. The CIM is trained on a set ${\mathcal{D}_K \coloneqq \bigcup^K_{\kappa =1}\mathcal{D}_{\kappa}^{\mathcal{T}}}$, where each ${\mathcal{D}^{\mathcal{T}}_{\kappa}} = \mathcal{\{X}_{\kappa}^{\mathcal{T}}, \kappa \}$ is a dataset of images corrupted with some corruption type $\kappa$,
and $\kappa$ is known. 
Note that $\kappa$ refers here to either endogenous or exogenous corruptions, depending on the available  data.}
CIM is trained end-to-end following \cite{millerclass} via distance-based contrastive training using a Class Anchor Clustering (CAC) loss  defined by

\begin{equation}
    \mathcal{L}_{CAC}(\mathbf{x}, y) = \mathcal{L}_{T}(\mathbf{x}, y) + \lambda \mathcal{L}_{A}(\mathbf{x}, y),
\end{equation}
where $\mathbf{x}$ is the input image with its label $y$ and $\lambda$ is a hyperparameter. The CAC loss aggregates two individual losses: i) a tuplet loss $\mathcal{L}_{T}(\mathbf{x}, y)$ \cite{tupletloss,millerclass} used to minimize the distance between training samples and their ground-truth anchored class centre, and ii) an anchor loss $\mathcal{L}_{A}(\mathbf{x}, y)$ \cite{millerclass} used to maximize the distance to other anchored class centres. Thereby, the CAC loss $\mathcal{L}_{CAC}$ encourages training data to form tight and class-specific clusters, and anchored class centres to fix cluster centre positions during training.

\textbf{Inference:}
After training the CIM model $r(\cdot)$ on $\mathcal{D}_K$,
the final layer $l(\cdot)$ is removed and the feature extractor ${g}(\cdot)$ 
is used to extract $q$-dimensional features ${\mathbf{z} \in \mathbb{R}^{q}}$ 
from corrupted samples\footnote{Empirically, we found that using the features $\mathbf{z}$ instead of corruption identification probabilities generalizes better to unseen corrupted test data.}.
Then, prototypical features ${\bar{\mathbf{z}}_{\kappa} = \frac{1}{h_\kappa} \sum_{i = 0}^{h_{\kappa}} \mathbf{z}_{i}}$ are computed from the training set, where each $\mathbf{z}_{i}$ is a feature vector corresponding to an image corrupted with corruption $\kappa$, and $h_{\kappa}$ is the number of samples affected by the corruption ${\kappa}$. 
The calculated ${K}$ prototypes are concatenated by {$\bar{\mathbf{Z}} = [ \bar{\mathbf{z}}_1^T, \bar{\mathbf{z}}_2^T, \ldots, \bar{\mathbf{z}}_K^T]^T$} to construct the prototype matrix $\bar{\mathbf{Z}}\in \mathbb{R}^{K \times q}$ where $(\cdot)^T$ denotes the vector/matrix transpose.

We employ a distance-based classifier $\phi(\cdot, \cdot)$ to classify features according to their relative distance to prototypical features. 
The classifier {{$\phi(\mathbf{z}, \bar{\mathbf{Z}})$ outputs ${\mathbf{d} = (||\mathbf{z}-\bar{\mathbf{z}}_1||_2, \dots, ||\mathbf{z}-\bar{\mathbf{z}}_K||_2)^{T}}$}} where $|| \cdot ||_2$ denotes the Euclidean norm. 

The output is normalized by ${\mathbf{b} = \mathbf{d} \ \odot \ (1-\text{softmin}(\mathbf{d}))}$ \cite{millerclass}, where $\odot$ is the element-wise product, and
\begin{equation} \small
\text{softmin}(\mathbf{d})_{\kappa} = \frac{\exp^{-{d}_{\kappa}}}{\sum_{{\kappa}=1}^{K}{\exp^{-{d}_{\kappa}}}}, \quad \mathbf{d} = [d_{\kappa}]_{{\kappa}=1}^K,
\end{equation}
is utilized to match the feature with the closest prototype.
Then, the model $r' = \phi \circ g$ predicts the corruption affecting the input by 
\begin{equation}
    \hat{\kappa} = \argmin_{\kappa}(\mathbf{b}).
   \label{eqk}
\end{equation}

\subsection{Per-corruption Adaptation of BN Statistics}
\label{sec:specialization}

{{\textbf{Batch Normalization (BN)} \cite{Krizhevsky2009LearningML} is a technique, used to make training of artificial neural networks faster and more stable through normalization of the layer inputs by re-centering and re-scaling. It is widely used in DNNs to mitigate the problem of internal covariate shift, where changes in the distribution of the inputs of each layer affect the learning of the network.
BN is applied over a 4D input (a mini-batch of 2D inputs with additional channel dimension) \cite{ioffe2015batch}.}}

Let ${\mathcal{B}}$ denote a mini-batch of features, obtained using model $F(\cdot)$, and let ${\mathbf{f} \in \mathcal{B} \subset \mathbb{R}^{B \times D \times L}}$ be a feature map in the mini-batch, {where $B$, $D$, and $L$ denote the batch size, the depth and the size of each feature map, respectively}. {{The mean $\mathbf{\mu} \in \mathbb{R}^D$ and standard-deviation $\mathbf{\sigma} \in \mathbb{R}^{D}$ (BN statistics) are employed per-dimension over the mini-batches channel-wise for normalizing features using}}
\begin{equation} \small
 \text{BN}(\mathbf{f}; \mathbf{\mu}, \mathbf{\sigma}^2) := \gamma \frac{\mathbf{f} - \mathbf{\mu}}{\sqrt{\mathbf{\sigma}^2 + \epsilon}} + \beta,
\end{equation}
where $\gamma$ and $\beta$ are learnable affine parameter vectors of size $D$, and $\epsilon > 0$ is a small constant used for numerical stability.

\textbf{Test-Time Adaptation (TTA)} refers to adapting DNNs to distribution shifts, with access to only the unlabelled test samples belonging to the target domain $\mathcal{T}$ at test time.
The conventional way of employing BN in test time is to set $\mu$ and $\sigma^2$ as those estimated from source data.
Instead, TTA methods estimate BN statistics directly from test batches to reduce the distribution shift at test time by
\begin{equation} \small \label{eq:bnupdate}
    {\mathbf{\mu}} = \frac{1}{{B\cdot L}} \sum_{\mathbf{f} \in \mathcal{B}}{\mathbf{f}}, \quad {\mathbf{\sigma}}^{2} = \frac{1}{{B\cdot L}} \sum_{\mathbf{f} \in \mathcal{B}}{(\mathbf{f} - {\mathbf{\mu}} )^2}.
\end{equation}
This practice is simple yet effective and thus adopted in many recent TTA studies \cite{Nado2020EvaluatingPB,Schneider2020ImprovingRA,wang2021tent,wang2022continual,gong2022note}. 
In our paper, we propose updating BN statistics via TTA, separately, per each corruption type, as described next.

\textbf{Estimating statistics on test data.}
Let $\Lambda \coloneqq ({\mu}, {\sigma}^2) \subset \mathcal{W}$ be the set of BN statistics of the model $F(\mathbf{x}, y; \mathcal{W})$. We denote the set of BN statistics obtained after training the model on the source dataset by $\Lambda^{\mathcal{S}}$.
We first initialize $K$ sets of source BN parameters $\Lambda^{\mathcal{S}}$. Then, we update each set according to the corruption type present in the input image.
In the ideal case, each set is associated to a specific corruption type $\kappa$, and each corruption type is always identified correctly. Therefore, the BN statistics $\Lambda_{\kappa}^{\mathcal{T}}$ associated with the type $\kappa$ are updated only with images corrupted with corruption type $\kappa$ that belong to the test set $\mathcal{D}_{\kappa}^{\mathcal{T}}$.
{\rm {We define this ideal reference set of statistics by $\Lambda_{\kappa}^\mathrm{ref}$}}. 
However, the target corrupted test images come without the corruption label $\kappa$, and BN parameters must be computed on the corruption type estimated by CIM ($\hat{\kappa})$. 

\subsection{Use CIM and TTA to Improve Task Performance}
\label{sec:codebook}
When deployed on a robotic device, our system is composed of (i) a CIM module (Sec.~\ref{sec:corrid}) employed to recognise the corruption type affecting the unlabelled input test image, and  (ii) $K$ sets of clean BN statistics, obtained training a model $F(\cdot)$ on clean training data (Sec.~\ref{sec:specialization}).
The purpose of our PAN is to improve the downstream task performance of $F(\cdot)$ by using CIM to identify the correct corruption type, update the correct set of BN parameters via TTA, and finally plug the updated set of BN parameters into the network.

\textbf{Codebook mapping.}
In detail, at inference time, for each input test image $\mathbf{x} \in \mathcal{X}^\mathcal{T}$, we estimate the corruption type using the CIM by $r'(\mathbf{x}) = \hat{\kappa}$. Then, we use a \textit{codebook} $\mathfrak{C}$ to map each estimated corruption type $\hat{\kappa}$ to a corruption-specific set of BN statistics $\Lambda_{\kappa}^{\mathcal{T}}$ by
\begin{equation}\label{eqcb}
    \mathfrak{C}: \hat{\kappa} \longmapsto \Lambda_{\hat{\kappa}}^{\mathcal{T}} :=(\mathbf{\gamma}_{\hat{\kappa}}, \mathbf{\beta}_{\hat{\kappa}}).
\end{equation}

Note that BN statistics associated with each of the $K$ corruptions are initialized as $\Lambda^{\mathcal{S}}$, and will be assigned to $\Lambda_{\hat{\kappa}}^{\mathcal{T}}$ after they are estimated by TTA.
The more CIM is able to correctly recognize the corruption (when $\hat{\kappa} = \kappa$), the more the BN statistics are specialized for such corruption and are different from the others. Fig.~\ref{fig:bn_layers} shows that our PAN can obtain BN statistics close to reference ones $\Lambda^{\mathcal{T}} \approx \Lambda^\mathrm{ref}$.

\begin{table*}[t]
    \centering \scriptsize
        \caption{\scriptsize PAN improves over different architectures, datasets and tasks. BN parameters' size is negligible with respect to the size of the whole model. 
        The number of parameters of each architecture is also reported with the percentage used for test-time adaptation. Real: natural corruptions.} \vskip -2ex
            \setlength{\tabcolsep}{3.5pt}
        \renewcommand{\arraystretch}{0.9}
    \begin{tabular}{lccccc ccc ccc}
    \toprule \scriptsize
        \textbf{Model} & \textbf{Backbone} & \textbf{Dataset} & \textbf{Task} &  \textbf{Real} & \textbf{Source} & \textbf{PAN (ours)} & \textbf{Gain (\%)} & \textbf{Metric} & \textbf{Model (MB)} & \textbf{BN (MB)} \\
        \cmidrule(lr){1-5} 
        \cmidrule(lr){6-9}
        \cmidrule(lr){10-11}
        ResNet18 \cite{resnet} & | & ImageNet-C \cite{hendrycks2019robustness} & Object Recognition & ~ & 31.7 & 39.0 & 23.0 & CA $\uparrow$ & 44.6 & 0.04 \\
        ResNet50 \cite{resnet} & | & ImageNet-C \cite{hendrycks2019robustness} & Object Recognition & ~ & 46.1 & 47.7 & 3.5 & CA $\uparrow$ & 97.7 & 0.20 \\
        ResNet101 \cite{resnet} & | & ImageNet-C \cite{hendrycks2019robustness} & Object Recognition & ~ & 53.0 & 55.5 & 4.7 & CA $\uparrow$ & 170.3 & 0.40 \\
        MobileNetV3 \cite{howard2017mobilenets} & | & ImageNet-C \cite{hendrycks2019robustness} & Object Recognition & ~ & 32.9 & 34.4 & 4.6 & CA $\uparrow$ & 9.7 & 0.05 \\
        ResNeXt50 \cite{resnext} & | & ImageNet-C \cite{hendrycks2019robustness} & Object Recognition & ~ & 49.6 & 51.3 & 3.4 & CA $\uparrow$ & 95.7 & 0.26 \\
        Wide-ResNet50 \cite{zagoruyko2017wide} & | & ImageNet-C \cite{hendrycks2019robustness} & Object Recognition & ~ & 49.0 & 50.2 & 2.4 & CA $\uparrow$ & 263.0 & 0.26 \\
         ResNet50 \cite{resnet} & | & VizWiz \cite{Bafghi_2023_CVPR} & Object Recognition & $\checkmark$ & 39.1 & 43.8 & 12.0 & CA $\uparrow$ & 97.7 & 0.20 \\
        ResNet50 \cite{resnet} & | & OpenLORIS \cite{she2019openlorisobject} & Object Recognition & $\checkmark$ & 42.5 & 43.8 & 3.1 & CA $\uparrow$ & 97.7 & 0.20 \\
        \midrule
        YOLOv8n \cite{yolov8_ultralytics} & CSPNet \cite{wang2019cspnet} & VOC-C \cite{michaelis2020benchmarking} & Object Detection & ~ & 34.6 & 36.3 & 4.9 & mAP$^{50-95}$ $\uparrow$ & 12.1 & 0.04 \\
        YOLOv8n \cite{yolov8_ultralytics} & CSPNet \cite{wang2019cspnet} & {ExDARK} \cite{Exdark} & Object Detection & $\checkmark$ & 39.4 & 40.3 & 2.3 & mAP$^{50-95}$ $\uparrow$ & 12.1 & 0.04 \\
        \midrule
        DeepLabV2 \cite{deeplab2_2021} & MobileNetV2 \cite{mobv2} & Cityscapes-C \cite{michaelis2020benchmarking} & Sem. Segmentation & ~ & 34.5 & 41.5 & 20.3 & mIoU $\uparrow$ & 42.2 & 0.11  \\
         DeepLabV2 \cite{deeplab2_2021} & MobileNetV2 \cite{mobv2} & ACDC \cite{SDV21} & Sem. Segmentation & $\checkmark$ & 37.8 & 40.1 & 6.1 & mIoU $\uparrow$ & 42.2 & 0.11 \\
        \bottomrule
    \end{tabular} \vspace{-1em}
    \label{tab:nets}
\end{table*}

\begin{figure}[t]
    \centering
    \includegraphics[width=\linewidth]{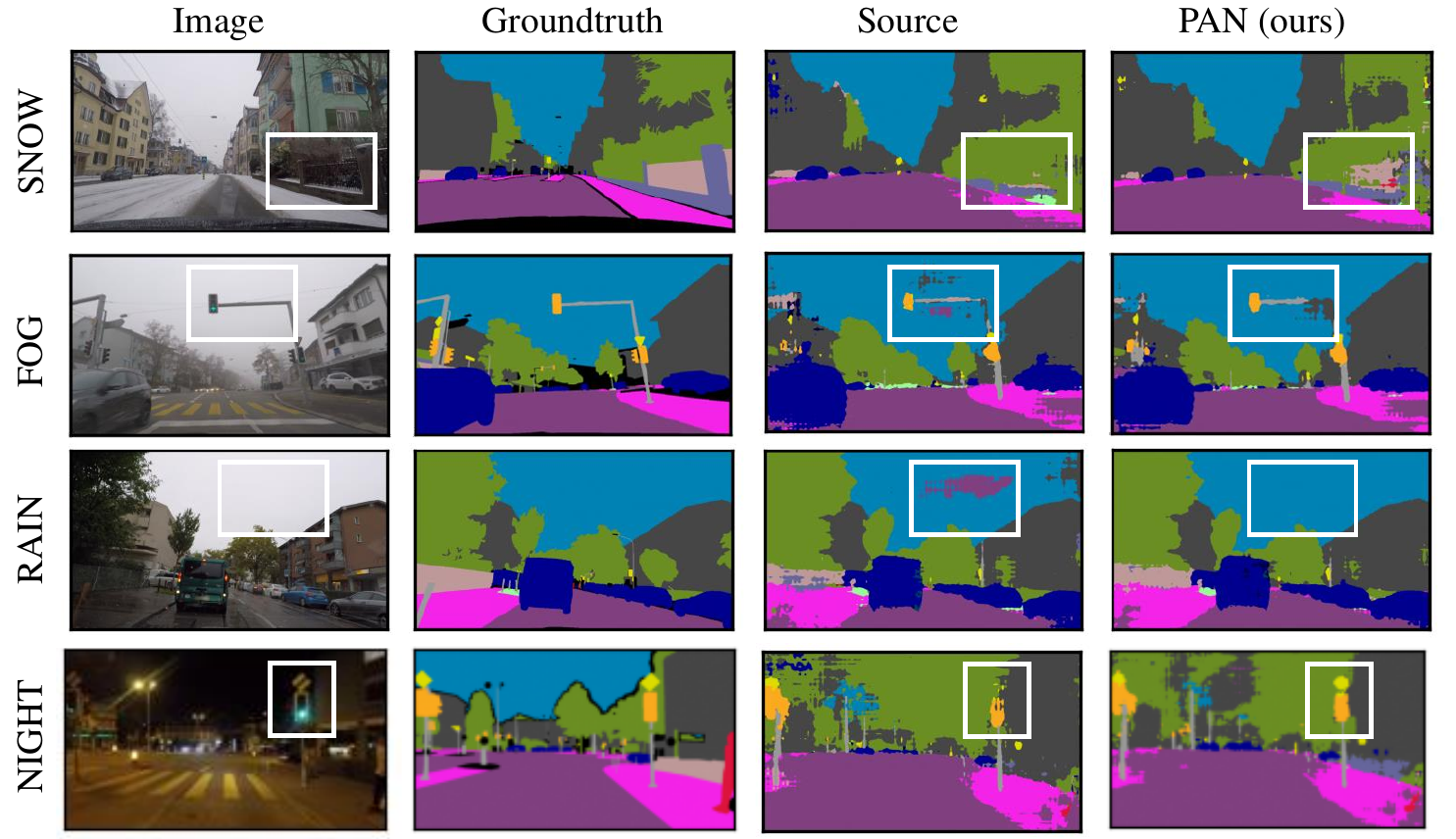}\vskip -1ex
    \caption{Per-Corruption qualitative results of semantic segmentation using the DeepLabV2 \cite{deeplab2_2021} with the ACDC \cite{SDV21} dataset.} 
    \label{fig:qual} \vspace{-1em}
\end{figure}

\section{Experiments}
\label{sec:results}

This section describes the experimental analysis of our approach.
We start our investigation by presenting results spanning different real and synthetic datasets, architectures, and visual scene understanding tasks, showing that PAN improves over the baseline. 
Then, we examine our approach on common synthetic recognition benchmarks by comparing our PAN against common methods used to solve dataset shift problems and delving into more detailed ablation studies.

As discussed throughout the paper, our PAN provides a versatile vision system, with limited memory and storage requirements that can be deployed on several robotic devices for different purposes (both indoors and outdoors).
For example, a robot vacuum cleaner equipped with PAN can navigate safely in low light conditions, and overexposed regions, and can promptly adapt to rapid illumination changes. 
In outdoor environments, we can appreciate the effects of PAN in \eg, autonomous vehicles: PAN is beneficial when navigating dark environments or in the presence of extreme weather conditions, such as fog, rain, or snow.

\textbf{Evaluation metrics} follow recent works \cite{hendrycks2019robustness,yucel2023hybridaugment,gong2022note,wang2021tent}. 
Robustness to corruptions is evaluated either using the Classification Accuracy (CA \%, $\uparrow$) or the mean Corruption Error (mCE, $\downarrow$), for the object recognition task. mCE calculates the average classification error
of the model on each corruption type over all the severity
levels, normalized by the error obtained using AlexNet \cite{hendrycks2019robustness}. We use mean Intersection over Union (mIoU \%, $\uparrow$) for semantic segmentation and mean Average Precision \cite{yolov8_ultralytics} (mAP$^{50-95}$ \%, $\uparrow$) for object detection. 

\subsection{Versatility of PAN on Robotic Vision Data} 

We present results on various robotics datasets, employing different models and performing different tasks, in order to show the versatility of PAN to outperform source models in various setups (Tab.~\ref{tab:nets}).
Following most of the recent approaches \cite{Khurana2021SITASI,yucel2023hybridaugment,hendrycks2019robustness}, we start showing experiments on datasets synthetically corrupted by exogenous distortions. 

First, we experiment on the \textbf{ImageNet-C} \cite{hendrycks2019robustness} using several different convolutional architectures (with BN layers). We show that the CA of the model grows by up to $7.2\%$ applying PAN on top of the source model, and the relative gain does not depend on the model architecture.
Second, we examine the applicability of PAN to various vision tasks, \ie, object recognition, detection and semantic segmentation with either synthetic or real-world corruptions.
Hence, we use the \textbf{VOC-C} \cite{michaelis2020benchmarking} for object detection and the \textbf{Cityscapes-C} \cite{michaelis2020benchmarking} for semantic segmentation (both with synthetic corruptions).

PAN improves accuracy significantly when applied to synthetically corrupted datasets on all tasks.  
{{We note that the percentage gain of PAN depends on the utilised model, but it is not much affected by the number of the employed BN layers (that is the only part of the architecture affected by PAN, as we will see in Sec.~\ref{ssec:abl}).}}
Despite obtaining good performance on synthetic distortions, many state-of-the-art technologies fail when applied to real-world corruptions. On the other hand, achieving good performance on these datasets is essential to deploy reliable AI systems on real robotic devices.
With this aim, we devise experiments training PAN on the VizWiz-classification \cite{Bafghi_2023_CVPR} dataset (detailed in the next subsection) and three common robotic vision benchmarks:

\textbf{OpenLORIS} \cite{she2019openlorisobject}. Object Recognition Dataset (OpenLORIS-Object) is designed for incremental recognition of common objects in office or home scenarios. The dataset includes some of the common challenges that home robots usually face, \eg, different illumination conditions, occlusions, camera-object distances/angles, and context information (clutters).\footnote{\label{note1}
To fit our purpose, we use the data with the lowest severity level for training and data with higher severity levels for testing.}

\textbf{ExDARK} \cite{Exdark}. The Exclusively Dark dataset is a collection of 7,363 low-light images captured in 10 different conditions from very low-light environments to twilight with 12 object classes annotated with object bounding boxes. The dataset is designed for indoor and outdoor \textit{object detection}.\footref{note1}

\textbf{ACDC} \cite{SDV20}. The Adverse Conditions Dataset with Correspondences is a popular benchmark in autonomous driving, generally used for \textit{semantic segmentation} in adverse visual conditions. It comprises a large set of 4006 images which are evenly distributed between fog, nighttime, rain, and snow.

We note that PAN outperforms the source model also in these challenging situations, even if the gain is smaller than those obtained on synthetic corruptions due to the less neat separation of corruption types applied to the input (\eg, multiple corruption types co-exist on target data).
Fig.~\ref{fig:qual} shows some qualitative results of PAN obtained by performing semantic segmentation in an outdoor autonomous driving environment. 
PAN shows better segmentation maps with respect to baseline predictions for every corruption. 
Tab.~\ref{tab:nets} reports also the model size and the size of the BN layers for each architecture used; the latter is small and almost negligible compared to the former.

\begin{table}[t]
    \centering\scriptsize    
    \caption{ \scriptsize CA (\%, $\uparrow$) on \textbf{CIFAR-C} with ResNet18. Comparisons from \cite{gong2022note}.}  \vskip -2ex
            \setlength{\tabcolsep}{4.1pt}
        \renewcommand{\arraystretch}{0.9}
    \begin{tabular}{l cc c}
    \toprule
         & \textbf{CIFAR10-C} & \textbf{CIFAR100-C} & \textbf{Avg CA} \\ 
          \cmidrule(lr){2-3}\cmidrule(lr){4-4}
        Source & 57.7 & 33.4 & 45.6 \\
        BN Adapt \cite{Schneider2020ImprovingRA} & 26.6 & 35.0 & 30.8 \\ 
        ONDA \cite{onda} & 36.4 & 50.4 & 43.4 \\ 
        Pseudo label \cite{Lee2013PseudoLabelT} & 24.6 & 33.6 & 29.1 \\ 
        TENT \cite{wang2021tent} & 23.6 & 33.1 & 28.3  \\ 
        CoTTA \cite{wang2022continual} & 24.5 & 35.8 & 30.1 \\ 
        LAME \cite{lame} & 63.9 & 36.7 & 50.3 \\ 
        NOTE \cite{gong2022note} & {78.9} & {53.0} & {65.9} \\ 
\midrule
\textbf{PAN (ours)} & \textbf{80.9}  & \textbf{62.1}  &
\textbf{71.5} \\ 
        \bottomrule
    \end{tabular}
    \label{tab:cifar}
\end{table}

\begin{table}[t]
    \centering \scriptsize
    \caption{\scriptsize CA (\%, $\uparrow$) on \textbf{VizWiz} \cite{Bafghi_2023_CVPR} with ResNet50. Left to right: 6 specific corruptions (\textit{blur}, \textit{bright}, \textit{framing}, \textit{rotation}, \textit{obscured}, \textit{dark}), average corrupted, clean source and total average; as defined in \cite{Bafghi_2023_CVPR, gurari2018vizwiz}.} \vskip -2ex
            \setlength{\tabcolsep}{3pt}
        \renewcommand{\arraystretch}{0.9}
    \begin{tabular}{l cccccc cc c}
    \toprule
        ~ & \textbf{BLR} & \textbf{BRT} & \textbf{FRM} & \textbf{ROT} & \textbf{OBS} & \textbf{DRK} & \textbf{Corr} & \textbf{Clean} & \textbf{Total} \\ 
        \cmidrule(lr){2-7}\cmidrule(lr){8-9}\cmidrule{10-10}
        Source & 39.3 & 32.8 & 36.2 & 26.5 & 23.7 & 40.8 & 33.2 & 42.8 & 39.1 \\ 
        AugMix \cite{hendrycks2020augmix} & 41.0 & 34.4 & 39.3 & 31.1 & 26.6 & 42.5 & 35.8 & 46.3 & 42.0 \\ 
        APR \cite{Chen_2021_ICCV}  & 37.6 & 31.3 & 35.7 & 27.0 & 26.0 & 38.0 & 32.6 & 43.2 & 38.9 \\ 
        HA \cite{yucel2023hybridaugment} & 37.7 & 32.2 & 35.6 & 26.9 & 27.2 & 40.1 & 33.3 & 41.3 & 38.3 \\ 
        DA \cite{Hendrycks_2021_ICCV} & 39.3 & 33.8 & 38.1 & 30.9 & 30.2 & 40.8 & 35.5 & 45.6 & 41.1 \\ 
        BN Adapt \cite{Schneider2020ImprovingRA} & 25.2 & 21.3 & 23.1 & 18.5 & 20.7 & 24.4 & 22.2 & 32.9 & 27.7 \\
        \midrule
       \textbf{PAN (ours)} & \textbf{41.2} & \textbf{40.6} & \textbf{39.4} & \textbf{34.4} & \textbf{36.1} & \textbf{44.3} & \textbf{39.3} & \textbf{47.9} & \textbf{43.8} \\ 
       \bottomrule
    \end{tabular}
    \label{tab:vizwiz} \vspace{-1em}
\end{table}

\begin{table}[t]
    \centering \scriptsize
    \caption{\scriptsize mCE (\%, $\downarrow$) on \textbf{ImageNet-C} with ResNet50. We compare state-of-the-art \colorbox{LightCyan}{data augmentation} (results from \cite{yucel2023hybridaugment}) and \colorbox{LightYellow}{TTA} (the first three results are obtained from \cite{Khurana2021SITASI}) methods. $^\heartsuit$: pre-trained on samples corrupted with target corruptions. $^{\diamondsuit}$: pre-trained via DA+AugMix+HA.} \vskip -2ex
            \setlength{\tabcolsep}{4pt}
        \renewcommand{\arraystretch}{0.9}
    \begin{tabular}{l cccc c}
    \toprule
         & {\textbf{Noise}} & {\textbf{Blur}} & {\textbf{Weather}} & {\textbf{Digital}} & \textbf{mCE} \\
    \cmidrule(lr){2-5}\cmidrule(lr){6-6}
        Source & 80 & 84 & 77 & 82 & 80.6 \\ 
        \rowcolor{LightCyan} Source$^\heartsuit$  & 64 & 59 & 59 & 64 & 61.4 \\
        \rowcolor{LightCyan} Patch Uniform \cite{lopes2019improving} &  68 & 79 & 73 & 76 & 74.3 \\ 
        \rowcolor{LightCyan} AA \cite{autoaugment}  & 70 & 80 & 69 & 72 & 72.7 \\ 
        \rowcolor{LightCyan} Random AA \cite{autoaugment} & 71 & 82 & 73 & 77 & 76.1 \\ 
        \rowcolor{LightCyan} MBPool \cite{pmlr-v97-zhang19a}  & 74 & 79 & 67 & 74 & 73.4 \\ 
        \rowcolor{LightCyan} SIN \cite{Rusak2020ASW}  & 70 & 80 & 71 & 73 & 73.3 \\ 
        \rowcolor{LightCyan} AugMix \cite{hendrycks2020augmix} & 66 & 71 & 68 & 69 & 68.4 \\ 
        \rowcolor{LightCyan} APR \cite{Chen_2021_ICCV} & 65 & 77 & 61 & 71 & 68.9 \\ 
        \rowcolor{LightCyan} HA \cite{yucel2023hybridaugment} & 57 & 70 & 65 & 69 & 65.8 \\ 
        \rowcolor{LightCyan} PixMix \cite{Hendrycks2021PixMixDP} & 52 & 79 & 60 & 69 & 65.8 \\ 
        \rowcolor{LightCyan} DA \cite{Hendrycks_2021_ICCV} & 46 & 71 & 62 & 66 & 62.0 \\ 
        \rowcolor{LightYellow} PTN \cite{Nado2020EvaluatingPB} & 110 & 124 & 133 & 144 & 128.7 \\ 
        \rowcolor{LightYellow}BN Adapt \cite{Schneider2020ImprovingRA} & 70 & 79 & 62 & 74 & 70.9 \\ 
        \rowcolor{LightYellow}AugBN \cite{Khurana2021SITASI} & 69 & 77 & 61 & 72 & 69.8 \\ 
        \rowcolor{LightYellow}TENT \cite{wang2021tent}  & 104 & 103 & 84 & 88 & 94.3 \\ 
        \rowcolor{LightYellow}SAR \cite{niu2023towards} & 78 & 93 & 65 & 86 & 80.5 \\
        \rowcolor{LightYellow}EATA \cite{niu2022efficient} & 68 & 83 & 56 & 70 & 69.2 \\ 
        \textbf{PAN (ours)}  & 74 & 76 & 50 & 67 & 66.0 \\ 
        \midrule
        \rowcolor{LightCyan} Source$^{\diamondsuit}$ & 45 & 59 & 56 & 62 & 56.1 \\
        \textbf{PAN$^{\diamondsuit}$ (ours)} & 42 & 51 & 50 & 52 & {49.4} \\  
        \bottomrule
    \end{tabular}
    \label{tab:inc} \vspace{-1em}
\end{table}

\subsection{Comparisons with Other Approaches}
Following previous works, we compare our method with state-of-the-art approaches on synthetically corrupted test datasets.
We use the CIFAR10, CIFAR100 \cite{Krizhevsky2009LearningML} and ImageNet \cite{russakovsky2015imagenet} datasets. The CIFAR datasets comprise of $50,000$ $32\times32$ training images. The ImageNet contains around $1.2$M images belonging to $1000$ different classes. Evaluation is performed on the corrupted versions of these datasets' test splits, \ie, CIFAR10-C, CIFAR100-C, and ImageNet-C \cite{hendrycks2019robustness}.  For those datasets, corruptions are simulated for 4 categories (\textit{noise}, \textit{blur}, \textit{weather}, \textit{digital}) with $K=15$ corruption types, each with 5 severity levels.
For a fair comparison with the prior art \cite{hendrycks2019robustness,yucel2023hybridaugment,wang2021tent,gong2022note}, we adopt the ResNet18 \cite{resnet} for the CIFAR datasets, and the ResNet50 \cite{resnet} for the ImageNet. We use \textit{source} pre-trained weights obtained from \cite{torchvision}. 
 Results obtained using \textbf{CIFAR-C} are reported in Tab.~\ref{tab:cifar}, comparing our PAN against several state-of-the-art TTA methods.
 Our approach improves the best competitor (\ie, NOTE) by $2.5\%$ and $17.2\%$ CA on the CIFAR10-C and CIFAR100-C, respectively. 
Results obtained using the \textbf{ImageNet-C} are reported in Tab.~\ref{tab:inc}. 
First, we observe that the model pre-trained on clean data (\ie, Source) suffers from severe performance degradation. This degradation is attenuated by applying DA and TTA approaches. 
However, DA methods require re-training the model to enable robustness and train a single set of parameters for all corruption types, while TTA can be implemented a posteriori after having a trained model. PAN outperforms all compared TTA methods (by a significant margin) and most DA approaches. 
We observe that pre-training the whole model with the corruptions encountered in the test set (Source$^\heartsuit$) significantly improves the results, outperforming most corruption-agnostic data augmentation approaches.
Nevertheless, PAN obtains a larger gain. We include this experiment to motivate our choice of specialising BN layer parameters to each corruption separately.

\begin{figure*}[t]
\centering 
\subfloat[]{
\begin{minipage}[b]{.32\textwidth}
    \centering \footnotesize
    \includegraphics[trim=0.9cm 0.8cm 0.8cm 0.8cm,clip,width=\linewidth]{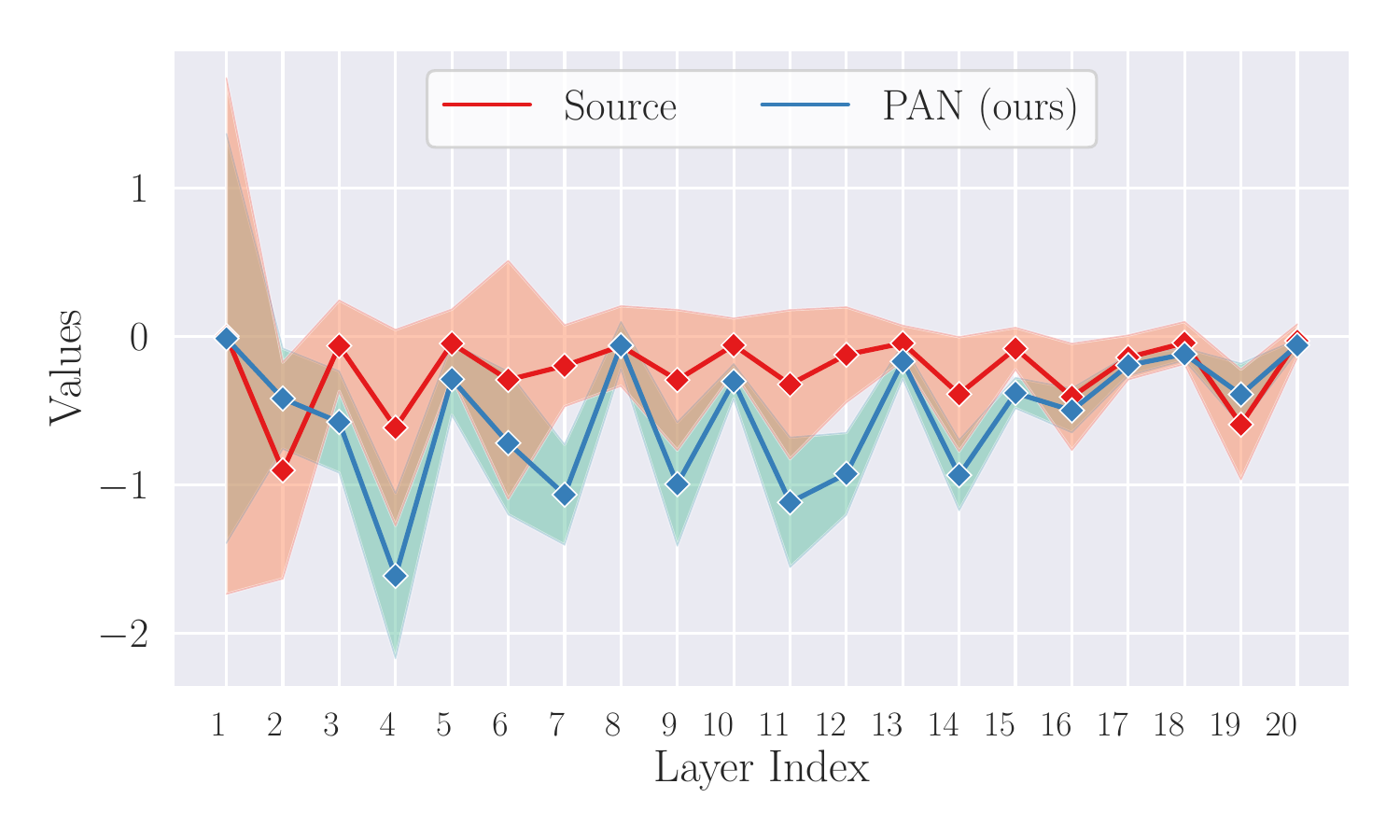} 
    \label{fig:per_layer} \vspace{-1em}
\end{minipage}
}
\subfloat[]{
\begin{minipage}[b]{.34\textwidth}
    \centering \footnotesize 
    \includegraphics[trim=0.5cm 1.7cm 0.3cm 0.8cm,clip,width=\linewidth]{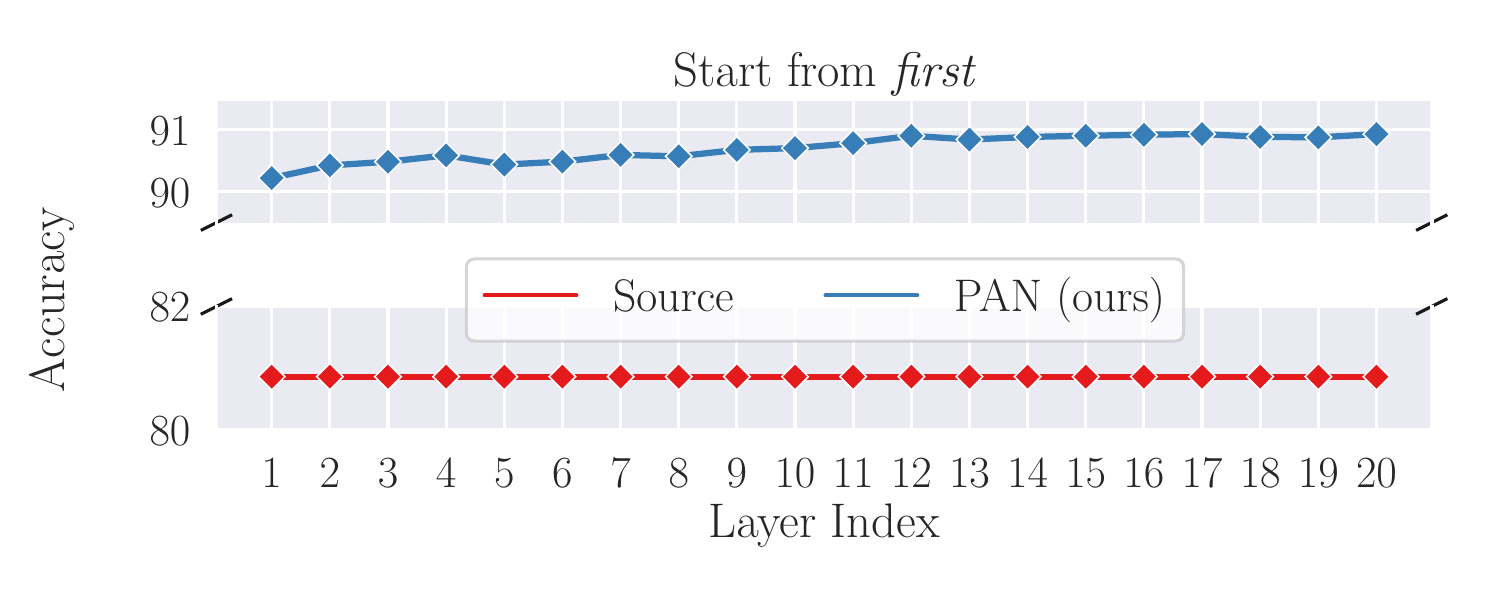}
    \includegraphics[trim=0.5cm 0.8cm 0cm 0cm,clip,width=\linewidth]{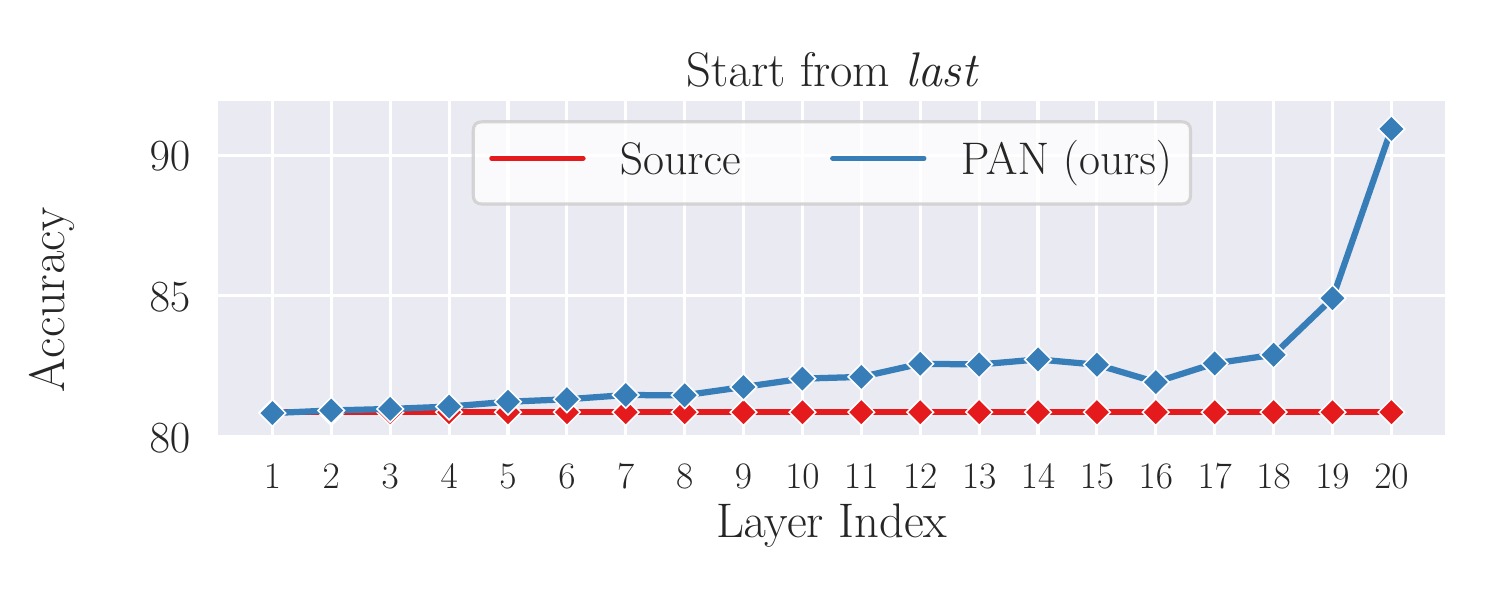} 
    \label{fig:bn_layers_first_last} \vspace{-1em}
\end{minipage}
}
\subfloat[]{
\begin{minipage}[b]{.32\textwidth}
    \centering \footnotesize 
    \includegraphics[trim=0.2cm 0.2cm 1.2cm 2cm,clip,width=\linewidth]{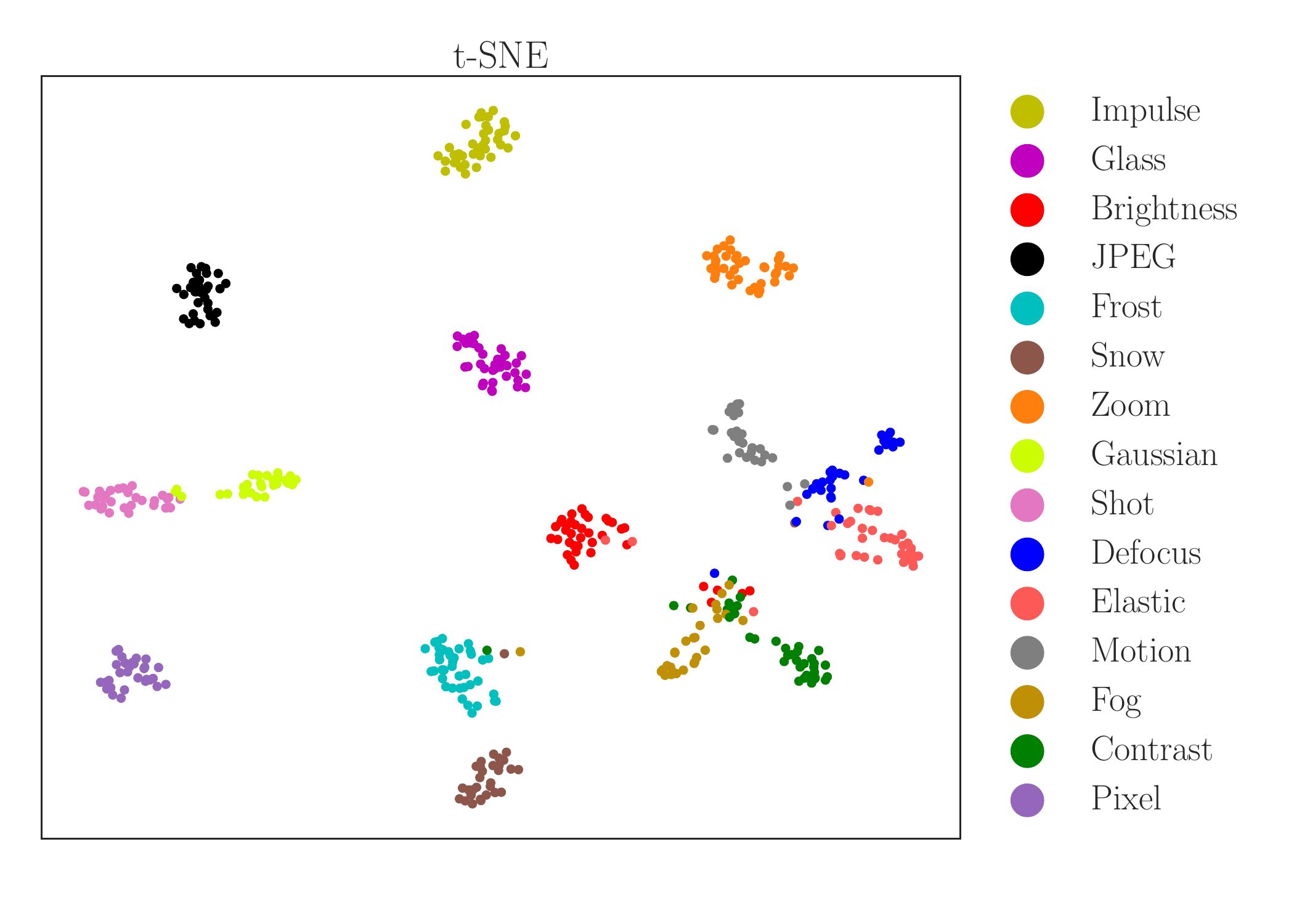}
    \label{fig:tsne} \vspace{-1em}
\end{minipage}}
\caption{\textbf{a:} Statistics of normalization layers are different throughout the network (e.g. for ResNet18 on CIFAR10-C). y-axis denotes the estimated mean $\mu$ values and the variance $\sigma^2$ is depicted by shadowed region. On average, per-corruption statistics identified by our approach (\textcolor{NavyBlue}{\textbf{blue}}) differ significantly from source ones (\textcolor{Red}{\textbf{red}}). {\quad\textbf{b:}} Change of CA for corrupted data for different normalization layers being adapted. We adapt all layers, {one at a time adding the next layer progressively}, starting from the first layer (\textbf{top}) or from the last layer (\textbf{bottom}). {\quad\textbf{c:}} t-SNE of per-corruption features produced by $g(\cdot)$. Different corruption types are clearly separated. Clusters of representations of similar corruption types are closer to each other (\eg, \textit{Shot} and \textit{Gaussian} noises; \textit{Fog} and \textit{Contrast}; \textit{Motion} and \textit{Defocus} blur).} \vspace{-1em}
\end{figure*}

Conversely, a robust model pre-trained with heavy data augmentation improves accuracy significantly over the standard Source baseline. 
The best augmentation approach is obtained by combining multiple state-of-the-art pipelines (\ie, Source$^{\diamondsuit}$ using DA+AugMix+HA).
TTA methods achieve comparable results with the key benefit of adaptation at test time rather than having to retrain the model from scratch with additional augmentations.
Our PAN can be seamlessly integrated on top of any pre-trained model. To examine this hypothesis, we employ PAN starting from pre-trained weights DA+AugMix+HA (PAN$^{\diamondsuit}$) and observe large gains by about $10\%$ relative mCE. %

Finally, we analyze some per-corruption results on the \textbf{VizWiz} \cite{Bafghi_2023_CVPR,chiu2020assessing} dataset in Tab.~\ref{tab:vizwiz}. The VizWiz is a recently proposed benchmark with images affected by natural corruptions. We use the classification split (built of $8,900$ images taken by blind people labelled with $200$ categories, \ie, a subset of the ImageNet label set) to test, and the other images with corruption labels for training. 
We observe that competing data augmentation and TTA methods that work well for synthetic corruptions fail or bring small improvements under this setting.  
PAN outperforms all the approaches on every per-corruption classification metric defined in \cite{Bafghi_2023_CVPR,chiu2020assessing}.
Moreover, by analyzing the results obtained on each corruption separately, we can note that the ROT images are the hardest to be correctly classified, and PAN improves CA by $7.9\%$. A large improvement occurs on BRT images, where we can observe a change in lighting condition similar to \textit{contrast} in the ImageNet-C dataset, where BN statistics were consistently shifted from the base ones  (Fig.~\ref{fig:hist}) implying a wider margin of improvement for PAN.
Overall, the contribution of PAN is major when the statistics of the BN layers of the network are shifted with respect to the source ones (this is the case of varying light/weather conditions, rotations and digital transformations) and minor elsewhere (\eg, on noisy or blurred samples).
These results suggest that PAN contributes to increasing the reliability of robot vision models: i) indoor robots can promptly adapt to instantaneous light changes and continue their navigation safely, and ii) outdoor autonomous vehicles can be more reliable in adverse weather conditions.
 
\subsection{Analyses and Ablation Studies}\label{ssec:abl}

\textbf{Per-corruption statistics} estimated using our approach are significantly different from those found using the model pre-trained on source domain data. 
We have analyzed how these statistics vary according to the input corruption type in Fig.~\ref{fig:bn_layers}.
Also, in Fig.~\ref{fig:per_layer}, we show that our method estimates diverse layer-wise normalization statistics with respect to the pre-trained source model. The highest divergence is shown in the first layers of the DNN model, as shown next.

\textbf{Selective adaptation} of statistics estimated at certain layers could be beneficial depending on the target hardware constraints. 
In this case, the most sensitive normalization parameters are those closer to the input, with the very first layer alone already accounting for $99.2\%$ of the total accuracy gain between the pre-trained source model and our PAN, as shown in the top plot of Fig.~\ref{fig:bn_layers_first_last}.
Conversely, adapting only the last normalization layers, we do not obtain large gains until we include the initial normalization layers (bottom plot of Fig.~\ref{fig:bn_layers_first_last}). 
This behaviour is expected as most of the variability on corrupted images affects the high-frequency components that are captured by the first layers of the network \cite{Sun2017ImprovingRO,yucel2023hybridaugment}.
{On the other hand, we accounted for all layers as considering them brings a bigger performance growth.}

\textbf{CIM's performance} is confirmed by Fig.~\ref{fig:tsne}, showing that the module is able to learn highly distinguishable features relative to different corruption types. Well-clustered features imply 
that ground truth corruption types are more easily recognized at inference time.  
{In terms of overhead, the CIM only adds a minimal $0.06$ ms inference time per image (on ImageNet-C), increasing the overall inference time of PAN from $1.54$ to $1.60$ ms on an NVIDIA GeForce RTX 2080 Ti.} %

\textbf{TTA performance} depends on the CIM.  Indeed, good performance of the CIM implies that each set of BN statistics is obtained performing TTA on a set of images, corrupted with the same corruption type. 
When the majority of images used to obtain $\Lambda_{\hat{\kappa}}^{\mathcal{T}}$ is corrupted with corruption type $\kappa$, we obtain a set of BN statistics $\Lambda_{\hat{\kappa}}^{\mathcal{T}}$ which results in being close to the set $\Lambda_{\kappa}^\mathrm{ref}$.
Hence, the normalization statistics identified by our \textit{codebook} $\mathfrak{C}$ are close to the reference BN statistics $\Lambda_{\kappa}^\mathrm{ref}$.
This can be noticed in Fig.~\ref{fig:bn_layers}, where we can also appreciate that PAN's BN statistics (\ie, $\Lambda_{\kappa}^{\mathcal{T}}$) are much closer to the per-corruption reference BN statistics compared to the pre-trained source ones (\ie, $\Lambda^{\mathcal{S}}$).

\section{Conclusion}
\label{sec:conclusion}

In this paper, we investigated the robustness of vision models in challenging environments, where acquired images are subject to several types of quality degradation. 
Our evaluation on images with natural distortions exposes the limitations of existing approaches that have mostly focused on synthetically corrupted data, emphasising the need for solutions to improve model robustness in practical real-world scenarios. 
Our method (PAN) identifies the corruption present in the target sample and uses such information to adapt batch normalization layers of downstream vision models to enhance their resilience.
PAN can be seamlessly plugged on top of any convolutional architecture employed for both indoor and outdoor robot systems, accomplishing many robot vision tasks (\eg, object recognition, detection, semantic segmentation).

\bibliographystyle{IEEEtran}
\bibliography{refs_short}

\end{document}